%% file: bare_jrnl_new_sample4.tex
\documentclass[lettersize,journal]{IEEEtran}
\usepackage{amsmath,amsfonts}
\usepackage{algorithmic}
\usepackage{algorithm}
\usepackage{array}
\usepackage[caption=false,font=normalsize,labelfont=sf,textfont=sf]{subfig}
\usepackage{textcomp}
\usepackage{stfloats}
\usepackage{url}
\usepackage{verbatim}
\usepackage{graphicx}
\usepackage{cite}

\usepackage{microtype}
\usepackage{bbding}

\usepackage{amsmath}
\usepackage{amssymb}
\usepackage{booktabs}
\usepackage{color}
\usepackage{colortbl}
\usepackage{graphicx}
\usepackage{makecell}
\usepackage{multicol}
\usepackage{multirow}
\usepackage{float} 

\begin{document}

\title{CVKD-UDA: Cross-View Knowledge Distillation for 3D Unsupervised Domain Adaptive Segmentation}

\author{Zhimin~Yuan,
        Ming~Cheng,~\IEEEmembership{Member,~IEEE,}
        Shangshu~Yu,
        Wen~Li, 
        Dunqiang~Liu, 
        Xin~Huang
        and~Cheng~Wang,~\IEEEmembership{Senior~Member,~IEEE}

        % <-this % stops a space
\thanks{Manuscript received 18 November 2025; revised 28 February 2026; accepted 1 May 2026. This work was supported in part by the Natural Science Foundation of Henan Province under Grant 262300422567, in part by the National Natural Science Foundation of China under Grant 62502243, in part by the Fundamental Research Funds for the Central Universities under Grant N25XQD053. \textit{(Corresponding author: Ming Cheng.)}}%

% \thanks{Manuscript received April 19, 2021; revised August 16, 2021. \textit{(Corresponding author: Ming Cheng.)}}

\thanks{Z. Yuan and X. Huang are with School of Artificial Intelligence, Nanyang Normal University, Nanyang 473061, China.}
\thanks{S. Yu is with the School of Computer Science and Engineering, Northeastern University, Shenyang 110819, China.}
\thanks{M. Cheng, W. Li, D. Liu and C. Wang are with Fujian Key Laboratory of Sensing and Computing for Smart Cities, School of Informatics, Xiamen University, Xiamen 361005, China (e-mail: chm99@xmu.edu.cn).}%
% \thanks{Digital Object Identifier ****}
}

\markboth{Submit to: IEEE Transactions on Multimedia
}%
{Yuan \MakeLowercase{\textit{et al.}}: \MakeUppercase{CVKD-UDA: Cross-View Knowledge Distillation for 3D Unsupervised Domain Adaptive Segmentation}}


\maketitle

\input{sections/00_abstract}
% \begin{abstract}
% This document describes the most common article elements and how to use the IEEEtran class with \LaTeX \ to produce files that are suitable for submission to the IEEE.  IEEEtran can produce conference, journal, and technical note (correspondence) papers with a suitable choice of class options. 
% \end{abstract}

\begin{IEEEkeywords}
Unsupervised domain adaptation, point cloud, cross-view knowledge distillation, semantic segmentation.
\end{IEEEkeywords}

\input{sections/01_intro}

\input{sections/02_related_work}

\input{sections/03_methods}

\input{sections/04_Experiments}

\input{sections/05_Conclusion}

\bibliographystyle{IEEEtran}

\bibliography{IEEEabrv, mybibfile}

\vfill

\end{document}

%% file: sections/00_abstract.tex
\begin{abstract}

3D unsupervised domain adaptive (UDA) segmentation mitigates the high cost of manual annotations of the new domain data. Self-training has emerged as the dominant approach in this area, where its success heavily depends on a well-initialized warm-up model to generate reliable pseudo labels. However, existing methods often depend on source supervision or output-level adversarial alignment to obtain the warm-up model, which suffer from limited generalization and training instability due to the large domain gap between domains. Constructing domain-similar representations is an effective way to bridge this gap. In this work, we propose CVKD-UDA, which revisits voxel size as a core design factor to construct domain-similar representations and leverages cross-view complementary cues to balance transferability and discriminability of the warm-up model. First, we generate two complementary views by varying voxel sizes and introduce a cross-view knowledge distillation (CVKD) to enhance generalization and target perception of the model. Second, to balance transferability and discriminability, we design a lightweight Decouple-Adapter and an auxiliary imitation classifier to decouple cross-view knowledge transfer. Extensive experiments on two benchmarks demonstrate that CVKD-UDA effectively improves the performance of self-training methods and provides a new perspective for 3D UDA segmentation. Our code will be available at GitHub.

\end{abstract}

%% file: sections/01_intro.tex
\section{Introduction}
\label{sec:intro}

\IEEEPARstart{S}{emantic} segmentation of 3D LiDAR point clouds is fundamental to achieving accurate spatial perception and comprehensive environmental understanding, facilitating applications in smart cities~\cite{park2019creating, xue2020lidar,guo2025neuv}, 3D map construction~\cite{mahajan2018construction}, autonomous driving~\cite{zhu2021cylindrical,ando2023rangevit}, etc. With the rapid development of supervised 3D segmentation methods~\cite{hu2020randla,ando2023rangevit,puy2023using,lai2022stratified,zhu2021cylindrical,zhao2023lif}, accurate segmentation results have been achieved. However, these methods require massive annotated data, which is costly and laborious to obtain, especially in complex urban driving scenarios. To alleviate this limitation, unsupervised domain adaptation~\cite{ben2010theory} (UDA) offers a promising solution, which transfers the knowledge learned from easily obtained synthetic data with rich annotations to real-world unlabeled data. 

Self-training (ST), which iteratively refines the model using pseudo labels generated from its own predictions, has become a dominant paradigm for 3D UDA segmentation~\cite{yuan2024density,saltori2022cosmix,liu2025sf,xiao2022polarmix,xiao2023domain}. Despite its promising performance, its success heavily depends on a well-initialized \textit{warm-up} (pre-trained) model to generate high-quality pseudo labels. Currently, the warm-up model adopted by most existing ST-based methods~\cite{saltori2022cosmix,xiao2022polarmix,xiao2023domain} is trained on the source domain in a fully supervised manner. However, the significant domain gap between source and target domains often results in noisy pseudo-labels that degrade the performance. Drawing inspiration from 2D UDA counterparts~\cite{zhang2021prototypical,shen2023diga}, the recent work DGT-ST~\cite{yuan2024density} adopts a two-stage training pipeline. It first conducts feature alignment via adversarial learning to obtain a warm-up model, and then refines the model through self-training. This approach achieves remarkable improvements, highlighting the critical role of a well-initialized warm-up model. Nevertheless, adversarial training often suffers from unstable optimization, hyperparameter sensitivity, etc. These drawbacks motivate us to design a simple yet effective method to obtain a well-initialized warm-up model, thereby facilitating subsequent self-training.

According to the theory of domain adaptation~\cite{ben2010theory,chen2019transferability}, two essential properties are required of the features learned by the model: 1) \textit{Transferability}, which allows the knowledge learned from the source domain can be effectively generalized to the target domain; 2) \textit{Discriminability}, which ensures that features are easily separable by the classifier. A common approach to improving transferability is to construct domain-similar representations that alleviate the domain gap. To this end, existing methods~\cite{xiao2022transfer, li2023adversarially, yuan2024density} often resort to generative adversarial networks (GAN) or region-wise statistical~\cite{yuan2024density} modeling to render source scans toward the target domain. This leads us to ask: \textit{Is there a simpler and more effective way to construct domain-similar representations?}

Currently, voxel-based methods are widely acknowledged as the dominant paradigm for 3D point clouds semantic segmentation, owing to their efficiency in processing large-scale 3D data. As shown in Fig.~\ref{fig:diff_voxs}, we visualize the voxelized representations of a LiDAR scan from different domains under varying voxel sizes. The results show that using a small voxel size retains fine-grained geometric details (standard view), which benefits discriminability but preserves significant domain differences. In contrast, moderately increasing the voxel size produces more structurally similar representations across domains (compressed view), which can be leveraged to improve the model's transferability. These observations highlight the critical role of voxel size in balancing transferability and discriminability. Therefore, rather than relying on adversarial training, we revisit voxel size as a core design factor to train a well-initialized warm-up model.

\input{figures/diff_voxs}

In this work, we propose a novel framework, CVKD-UDA, which is centered on a Cross-View Knowledge Distillation (CVKD) strategy. The core idea is to exploit compressed voxel representations to enhance cross-domain transferability while preserving discriminability in the standard view. Specifically, for transferability, CVKD performs cross-view knowledge distillation between two complementary voxelized representations generated with different voxel sizes. This label-free design allows the model to exploit both the source and target data simultaneously, enhancing its generalization capacity and perception of the target domain data. However, the introduction of an additional compressed view, which inevitably entails the loss of fine-grained geometric information, compromises the model’s discriminative capacity and consequently hinders accurate segmentation under the standard view. Thus, for discriminability, we propose a Decouple-Adapter (D-Adapter), an auxiliary imitation classifier, and an entropy regularization loss.
The two modules are designed to enhance the model’s flexibility, while the entropy regularization further prevents degenerate optimization during consistency training by encouraging confident representations. Notably, these modules are employed only during training and are discarded during inference. As the test phase requires only the standard view, our method introduces no extra computational overhead and can be seamlessly integrated into existing self-training frameworks.

Our main contributions can be summarized as follows:

\begin{itemize}
    
    \item We identify the critical role of voxel size in 3D UDA segmentation and propose CVKD-UDA, which leverages complementary views to balance transferability and discriminability for effective warm-up model training.
    
    \item To enhance transferability, we propose a cross-view knowledge distillation strategy that transfers knowledge between two complementary views. To preserve discriminability, we design a D-Adapter and an auxiliary imitation classifier that decouple cross-view learning and enable accurate segmentation under the standard view.

    \item Extensive experiments demonstrate that our warm-up model consistently boosts self-training performance and achieves superior results on two 3D UDA benchmarks, offering a new perspective for advancing research in 3D LiDAR UDA segmentation.

\end{itemize}

%% file: figures/diff_voxs.tex
\begin{figure*}[t]
\centering
\includegraphics[width=0.9\linewidth]{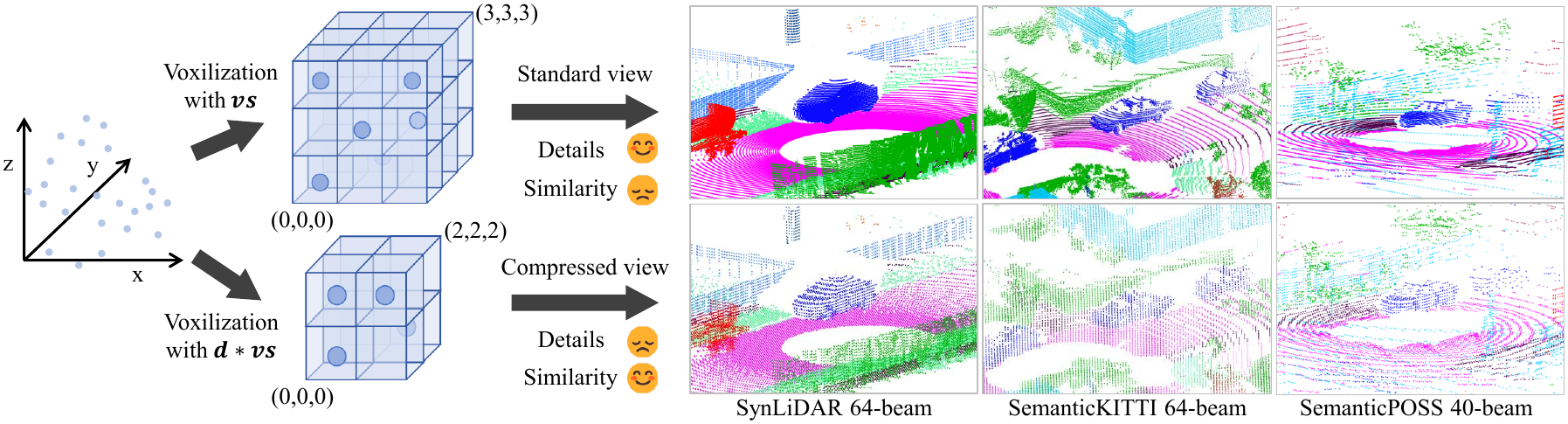}
\caption{Pipeline for generating two complementary views. Given a LiDAR scan from either the source or target domain, we voxelize it using two different voxel sizes to produce complementary representations for 3D UDA segmentation. The standard view, generated with a smaller voxel size, preserves fine-grained geometry but also retains larger inter-domain discrepancies (top). The compressed view, obtained with a moderately enlarged voxel size, smooths structural variations and enhances cross-domain similarity (bottom).}

\label{fig:diff_voxs}
\end{figure*}

%% file: sections/02_related_work.tex
\section{Related Work}
\label{sec:related_work}

\subsection{3D point clouds UDA segmentation} 

Unsupervised domain adaptive segmentation of 3D point clouds aims to leverage the labeled source and unlabeled target data to avoid the labor-intensive point-wise annotation in the new target domain. Existing methods can be roughly grouped into two categories: adversarial training and self-training.

For \textit{adversarial training}, many methods leverage generative adversarial network (GAN)~\cite{goodfellow2020generative} to construct domain-similar representations. ePointDA\cite{zhao2021epointda} adopts CycleGAN~\cite{zhu2017unpaired} to explicitly model dropout noise for domain alignment, while Complete $\&$ Label~\cite{yi2021complete} completes scans from both domains into a shared canonical space. PCT~\cite{xiao2022transfer} employs two GANs to transform appearance and sparsity separately, and ASM~\cite{li2023adversarially} learns a masking module to simulate target-specific noise patterns. Beyond input-level adaptation, several works~\cite{yuan2022category,yuan2023prototype,kreutz2024lion,yuan2024density} perform feature-level alignment via GAN to pretrain a warm-up model for self-training. However, this line of methods suffers from unstable training and sensitivity to hyperparameters, limiting their scalability to real-world applications.

In contrast, \textit{self-training}~\cite{michele2024saluda,saltori2022cosmix,xiao2022polarmix,shaban2023lidar,li2024construct} has emerged as a more scalable and effective alternative. These methods typically use the Mean-Teacher framework~\cite{tarvainen2017mean} to generate pseudo-labels and gradually adapt to the target domain. ConDA~\cite{kong2023conda} introduces a domain-concatenation strategy to exchange contextual signals; PolarMix~\cite{xiao2022polarmix} mixes point regions across domains to improve alignment, while CosMix~\cite{saltori2022cosmix} further incorporates semantic and structural cues. Despite their success, most approaches rely on a warm-up model trained on labeled source data, making performance sensitive to pseudo-label quality. To address this, several methods~\cite{gebrehiwot2023t,shaban2023lidar} adopt a two-stage pipeline. For example, SALUDA~\cite{michele2024saluda} learns an implicit surface representation as an auxiliary task on both domains to initialize a transferable model. LiDAR-UDA~\cite{shaban2023lidar} enhances source pretraining with subsampling and data augmentation. DGT-ST~\cite{yuan2024density} incorporates adversarial training during warm-up to boost transferability, followed by self-training to achieve further gains.

\textit{UDA for indoor 3D segmentation.} Besides the outdoor LiDAR UDA segmentation, a closely related direction is indoor synthetic-to-real UDA segmentation based on RGB-D reconstructed point clouds. DODA~\cite{ding2022doda} builds an intermediate domain by simulating realistic scanning artifacts and augmenting scene layouts to reduce both point-pattern (e.g., occlusion and noise) and context gaps. CACE~\cite{chen2025cace} further improves this idea with context-aware augmentations and prediction consistency. Algorithmically, these methods share a similar paradigm, which combines self-training with mix-based augmentation and can be adapted to the outdoor scenario. Nevertheless, as pointed out in DODA, directly reusing methods across 3D settings can be sub-optimal, since the dominant gap sources are different. Outdoor LiDAR shifts are mainly driven by sensor sampling patterns and physical factors (e.g., beam numbers, FoV, mounting, etc.), whereas indoor shifts largely arise from RGB-D reconstruction artifacts and layout/context discrepancies. Thus, indoor 3D UDA methods are informative for LiDAR adaptation, but they cannot be directly applied without accounting for LiDAR-specific sensing characteristics.

\subsection{Knowledge distillation} 

This technique~\cite{hinton2015distilling, liu2019structured} is originally proposed for model compression~\cite{bucilua2006model,sau2016deep,li2025adaptive}, where a lightweight student model learns to mimic a larger, well-trained teacher by aligning output predictions or intermediate features. Our work is inspired by DiGA~\cite{shen2023diga}, which introduces a distillation strategy that relies solely on source data to obtain a warm-up model. In contrast, we leverage the unique representational characteristics of LiDAR point clouds to construct domain-similar representations and use both source and target domain data to perform cross-view knowledge distillation.

%% file: sections/03_methods.tex
\section{Method}
\label{sec:methods}
\input{figures/CVKD}

For 3D point cloud UDA semantic segmentation, we have a labeled source domain $ \mathcal{D}_s = \{(x_s^{i}, y_s^{i})\}_{i=1}^{N_s} $ and an unlabeled target domain $\mathcal{D}_t = \{x_t^{i}\}_{i=1}^{N_t}$, where $N_s$ and $N_t$ are the number of scans of each domain. $y_s^{i} \in \{0, 1\}^{M_i \times K}$ represents the set of per-point one-hot labels for $M_i$ points over $K$ semantic classes. While these two domains share the same label space, their data distributions differ significantly due to variations in LiDAR configurations, environmental conditions, etc. We aim to learn a segmentation model that effectively transfers knowledge from $\mathcal{D}_s$ to $\mathcal{D}_t$, thereby obtaining accurate predictions on $\mathcal{D}_t$. To this end, we focus on improving transferability and discriminability of the model, which are explored in the following sub-sections.

\subsection{Cross-view Distillation for Transferability}

\textit{(1) Domain Similar Voxelized Representations.} Currently, the voxel-based methods are the dominant choice for 3D driving scenario point cloud segmentation. Given a point $\mathbf{p}_i = (x_i, y_i, z_i)$, its voxel representation is obtained by mapping the point to its discretized grid location as follows:
\begin{align}
V = ( 
\left\lfloor \frac{x_i - X_{\min}}{vs} \right\rfloor, \ 
\left\lfloor \frac{y_i - Y_{\min}}{vs} \right\rfloor, \ 
\left\lfloor \frac{z_i - Z_{\min}}{vs} \right\rfloor 
),
\end{align} 
where $V$ denotes the discretized voxel coordinate. $X_{\min}$, $Y_{\min}$, and $Z_{\min}$ denote the lower bounds of the voxel grid along each axis. $vs$ is voxel size, which typically adopts a small value (\textit{e.g}., $ 5 cm$) to preserve fine-grained geometry details for precise segmentation. However, such fine-grained voxelization also makes the model sensitive to point density variations, typically caused by differences in LiDAR configurations and environmental conditions between domains. Thus, the transferability of the model is limited.

To improve the transferability of the warm-up model, we explore constructing domain-similar voxel representations. We randomly choose a scan from different domains and visualize voxelized representations using varying voxel sizes in Fig.~\ref{fig:diff_voxs}. We observe that using a small voxel size $vs$ helps retain detailed geometric structures of the point cloud, but results in large domain discrepancies. In contrast, moderately enlarging the voxel size to $ d\cdot vs$  (where $d > 1$ is the voxel size expansion ratio) smooths out fine-grained variations and reduces the domain differences to some extent.

Building upon this observation, we construct two complementary views for each scan from the source and target domains:  
1) a standard view $x_{sv}$, voxelized with a smaller voxel size $vs$, which preserves fine-grained geometric structures to support accurate segmentation;  
2) a compressed view $x_{cv}$, voxelized with an moderately enlarged voxel size $ d \cdot vs $, which reduces domain-specific variations and enhances the transferability.  
We aim to jointly leverage these two views and design a framework to train a warm-up model with robust domain generalization while maintaining strong discriminative capability.
% In the following, w

\textit{(2) Cross-View Knowledge Distillation (CVKD).} To enhance the transferability of the warm-up model, we propose CVKD by leveraging the above-constructed $x_{sv}$ and $x_{cv}$. As shown in Fig.~\ref{fig:CVKD}, we adopt the Mean-Teacher framework and enforce consistency between $x_{sv}$ and $ x_{cv} $ in the output spaces. Notably, instead of aligning identical views from both the teacher and student models, we adopt a cross-view alignment strategy. The standard view predictions of the student model are aligned with the teacher's compressed view outputs to enhance transferability, and the teacher's standard view outputs guide the student's compressed view predictions to preserve discriminative capability. 

Specifically, for each scan from the source and target domains, we first concatenate $x_{sv}^{s/t}$ and $x_{cv}^{s/t}$, and feed them into the teacher and student models, respectively. The teacher and student models then produce predictions $\{p_{sv}^{s/t, tea/stu}, p_{cv}^{s/t, tea/stu}\}$ for both views, where the superscripts $s$ and $t$ indicate the source and target domains, respectively. We subsequently enforce cross-view consistency constraints at the output space to enable effective knowledge transfer between views. We employ the KL-divergence $D_{\mathrm{KL}}(\cdot \,\|\, \cdot)$ for output-level consistency:  
\begin{align}
\label{eq:logit_consist}
    \mathcal{L}_{logit}^{s} = D_{\mathrm{KL}}(p_{sv}^{s,stu} \,\|\, p_{cv}^{s,tea}) + D_{\mathrm{KL}}(p_{cv}^{s,stu} \,\|\, p_{sv}^{s,tea}).
\end{align}

Since CVKD is label-free and built on a self-distillation strategy, it can be seamlessly extended to the unlabeled target domain to enhance the warm-up model’s ability to perceive target domain information. For each unlabeled scan from the target domain, we employ the same two-view generation procedure as in the source and optimize the model by enforcing the consistency loss $\mathcal{L}^t_{logit}$:
\begin{align}
\label{eq:tgt_logit_consist}
    \mathcal{L}_{logit}^{t} = D_{\mathrm{KL}}(p_{sv}^{t,stu} \,\|\, p_{cv}^{t,tea}) + D_{\mathrm{KL}}(p_{cv}^{t,stu} \,\|\, p_{sv}^{t,tea}).
\end{align}

\subsection{Components for Preserving Discriminability}

% While CVKD improves the warm-up model's transferability, using an additional compressed view, while improving cross-domain generalization, inevitably weakens the model’s discriminative ability on the standard view. 

While CVKD improves the warm-up model's transferability and cross-domain generalization by using an additional compressed view, it inevitably weakens the model’s discriminative ability on the standard view. We empirically find that this compromises the model’s discriminative capacity and negatively impacts the performance of self-training, as detailed in ablation studies. To mitigate this, we propose two components and a regularization loss that enhance the model’s flexibility and enable robust representation learning across views. Notably, these modules are only used during training, without introducing extra computational cost during testing.

\textit{(1) Decouple-Adapter (D-Adapter).} Since final segmentation is conducted under $x_{sv}$ to produce fine-grained point-wise predictions, preserving discriminative features in $x_{sv}$ is critical. Sharing a single backbone to extract features from both $x_{sv}$ and $x_{cv}$ may cause feature entanglement, where the coarser geometry of $x_{cv}$ adversely affects the fine-grained representation learning required for accurate segmentation on $x_{sv}$. To address this, we draw inspiration from recent advances in parameter-efficient fine-tuning~\cite{chen2022vision,han2024parameter} and propose D-Adapter, a compact feature extraction module specifically designed for the compressed view. Given the feature $f_{cmp}$ of the compressed view extracted by the shared backbone, the D-Adapter refines it through a sequence of MLP, BatchNorm (BN), ReLU, and another MLP layer:
\begin{align}
\hat{f}_{cmp} = \text{MLP}_2(\text{ReLU}(\text{BN}(\text{MLP}_1(f_{cmp})))).
\end{align}
As shown in Fig.~\ref{fig:CVKD}, we place the D-Adapter before the final decode stage (\textit{i.e}., Block\_8 in MinkUnet34), allowing it to process high-level features of $x_{cv}$ before prediction. This module enables the model to extract view-specific features from $x_{cv}$ and reduce the interference on $x_{sv}$ representation, achieving a better balance between domain-invariant feature extraction and accurate segmentation.

\textit{(2) Auxiliary Imitation Classifier.} While the D-Adapter mitigates feature-level entanglement, $x_{cv}$ can still degrade the classifier’s discriminative capacity on $x_{sv}$. Directly enforcing output-level consistency between the student's predictions on $x_{sv}$ and the teacher's predictions on $x_{cv}$ (\textit{i.e}., matching their logits via $D_{\mathrm{KL}}(p_{sv}^{t,stu} \,\|\, p_{cv}^{t,tea})$) further amplifies this issue, limiting the main classifier to perform accurate predictions on $x_{sv}$. To address this issue, we propose two strategies. First, we upsample the compressed view features to match the resolution of the standard view. This enables the main classifier to focus on fine-grained semantic learning, partially mitigating interference from the compressed view. Second, we introduce an auxiliary imitation head into the student model to decouple CVKD from the main classifier. Instead of matching the outputs from the main classifier, we compute the KL-divergence between the teacher's main predictions ($p_{sv/cv}^{s/t,tea}$) and the outputs of the student’s auxiliary classifier ($z_{sv/cv}^{s/t,stu}$). This can preserve the discriminative capacity of the main classifier while still enabling effective cross-view knowledge transfer. Thus, the output-level consistency loss $\mathcal{L}_{logit}^{s}$ and $\mathcal{L}_{logit}^{t}$ which  originally defined in Eq.~\ref{eq:logit_consist} and Eq.~\ref{eq:tgt_logit_consist} are modified as:
\begin{align}
    \mathcal{L}_{logit}^{s} = D_{\mathrm{KL}}(z_{sv}^{s,stu} \,\|\, p_{cv}^{s,tea}) + D_{\mathrm{KL}}(z_{cv}^{s,stu} \,\|\, p_{sv}^{s,tea}),
\end{align} 
\begin{align}
    \label{eq:tgt_logit_consist_modified}
    \mathcal{L}_{logit}^{t} = D_{\mathrm{KL}}(z_{sv}^{t,stu} \,\|\, p_{cv}^{t,tea}) + D_{\mathrm{KL}}(z_{cv}^{t,stu} \,\|\, p_{sv}^{t,tea}).
\end{align} 

\textit{(3) Entropy Regularization.} The above modules are designed to enhance the flexibility of the model, while the optimization process remains prone to degeneration. Specifically, within the proposed consistency optimization framework, minimizing the KL-divergence loss may result in degenerate optimization, as the model can trivially reduce the consistency loss by generating uniform predictions across categories rather than learning discriminative representations.  For the labeled source domain, this issue is inherently mitigated because the supervision from ground-truth annotations provides a strong learning signal that regularizes the predictions (i.e., Cross-Entropy loss). 

However, for the unlabeled target domain, the model is optimized only through the KL-divergence Eq.~\ref{eq:tgt_logit_consist_modified}, rendering it vulnerable to degenerate solutions. This degenerate behavior impedes the learning of meaningful representations from the target domain and weakens the model’s ability to achieve effective adaptation. To mitigate this, we introduce an entropy regularization loss on the standard view of the target domain:
\begin{align}
\label{eq:tgt_ent_loss}
\mathcal{L}^{t, sv}_{ent}
= - \sum_{i}\sum_{c} p_{i,c}\log p_{i,c}
,
\end{align}
where $p_{i,c}$ denotes the prediction probability computed over voxels $i$ and classes $c$ of the $p^{t,\mathrm{stu}}_{sv}$. Since the compressed view distills knowledge from the standard view through CVKD, enforcing confident (low-entropy) predictions on the standard view implicitly guides the compressed view toward more discriminative outputs as well. This regularization term encourages the model to generate confident and well-separated predictions, effectively suppressing degenerate uniform outputs and promoting stable and discriminative target-domain learning. 

\subsection{Objective Functions}
Our training objective consists of two components: a consistency distillation loss and a segmentation loss. Specifically, to enhance transferability, we employ a cross-view distillation loss that enforces consistency in the output spaces across the two views, which is defined as:
\begin{align}
\mathcal{L}_{distill} = \lambda (\mathcal{L}_{logit}^{s} + \mathcal{L}_{logit}^{t}) + \beta \mathcal{L}_{ent}^{t}
\label{eq:distill}
\end{align}
where $\lambda$ and $\beta$ are two hyperparameters that control the relative weights of the logit consistency and entropy regularization.

In addition, to maintain accurate semantic prediction, we apply the cross-entropy loss $\mathcal{L}_{ce}^s$ to the standard and compressed views of the labeled source data:
\begin{align}
    \mathcal{L}_{seg}^{s} = \mathcal{L}_{ce,sv}^{s} + \mathcal{L}_{ce,cv}^{s},
\end{align}
\begin{align}
    \mathcal{L}_{ce}^{s} &= -\sum_{i=1}^{N_s}\sum_{c=1}^{K} y_{i,c}^s \log(p_{i,c}^{s}).
\end{align}

Thus, the final objective function $\mathcal{L}_{total}$ of the proposed CVKD-UDA is defined as:
\begin{align}
\label{eq:total_loss}
\mathcal{L}_{total} = \mathcal{L}_{seg}^{s} + \mathcal{L}_{distill}.
\end{align}

%% file: figures/CVKD.tex
\begin{figure*}[t]
\centering
\includegraphics[width=0.9\linewidth]{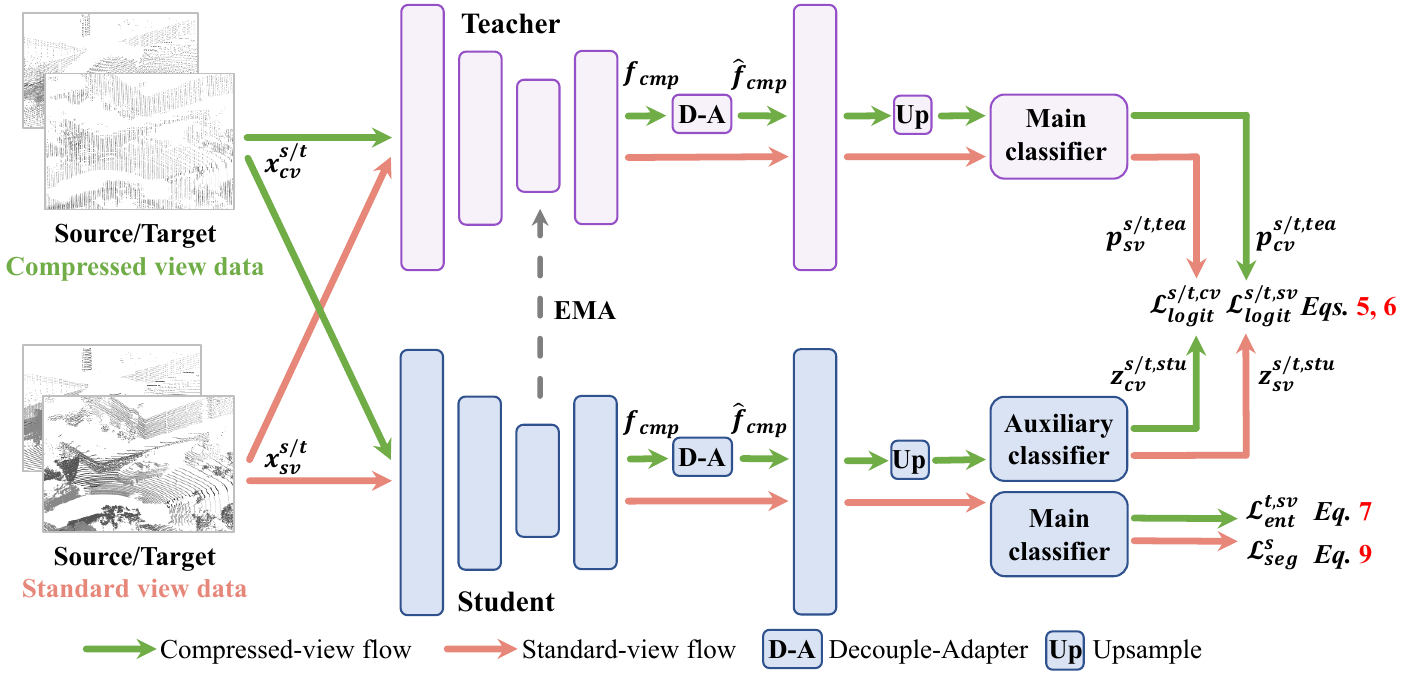}
\caption{Overview of the proposed CVKD-UDA. Given an input LiDAR scan from either the source or target domain, we generate two voxelized representations with different voxel sizes: a compressed view $x_{cv}$ and a standard view $x_{sv}$. These two views are concatenated and fed into a Mean-Teacher framework, where the teacher supervises the student via output-level cross-view distillation. D-Adapter is a lightweight module specifically designed to extract compressed view features. Furthermore, an auxiliary classifier is added to the student to decouple the knowledge transfer from the main classifier. }
\label{fig:CVKD}
\end{figure*}

%% file: sections/04_Experiments.tex
\section{Experiments}
\label{sec:exp}

\input{tables/syn2sk}

\subsection{Experimental Settings}
\textit{(1) Datasets.} Following the previous works~\cite{saltori2022cosmix,yuan2024density,yuan2023prototype}, we evaluate CVKD-UDA on two 3D UDA tasks: SynLiDAR $\rightarrow$ SemanticKITTI and SynLiDAR$ \rightarrow$ SemanticPOSS. SynLiDAR provides class mapping details to facilitate label matching.

% The official SynLiDAR~\cite{xiao2022transfer} repository provides the class mapping details to facilitate label matching for each domain adaptation task.

\label{subsec:exp_setting}

Specifically, \textit{SynLiDAR}~\cite{xiao2022transfer} is a large-scale synthetic dataset created using a LiDAR simulator that mimics the Velodyne HDL-64E sensor. It covers diverse virtual driving scenarios and comprises 13 sequences, totaling 19,840 annotated scans across 32 semantic categories. \textit{SemanticKITTI}~\cite{behley2019semantickitti} is a widely used real-world dataset collected in Germany by a Velodyne HDL-64E LiDAR. It provides point-wise annotations for 28 semantic categories. We use sequences 00–07 and 09–10 for training (19,130 scans), and reserve sequence 08 (4,071 scans) for validation. \textit{SemanticPOSS}~\cite{pan2020semanticposs} is a dataset collected at Peking University using a Pandora-40 sensor. It contains 6 sequences with point-wise annotations over 14 semantic categories. We use sequence 03 (500 scans) for validation, and the remaining 2,488 scans for training.

% \textit{SynLiDAR}~\cite{xiao2022transfer} is a large-scale synthetic dataset created using a LiDAR simulator that mimics the Velodyne HDL-64E sensor. In the virtual environment, the LiDAR sensor’s height, angle, and rotational speed are carefully configured to ensure a wide scanning range with rich geometric details, while maintaining a high degree of similarity to real-world data acquisitions. It covers diverse virtual driving scenarios and comprises 13 sequences, totaling 19,840 annotated scans across 32 semantic categories. 

% \textit{SemanticKITTI}~\cite{behley2019semantickitti} is a widely used real-world dataset collected in Germany by a Velodyne HDL-64E LiDAR. It provides point-wise annotations for 28 semantic categories. We use sequences 00–07 and 09–10 for training (19,130 scans) and reserve sequence 08 (4,071 scans) for validation of each adaptation approach. 

% \textit{SemanticPOSS}~\cite{pan2020semanticposs} is a dataset collected at Peking University using a Pandora-40 sensor. It contains 6 sequences with point-wise annotations over 14 semantic categories. We use sequence 03 (500 scans) for validation, and the remaining 2,488 scans for training.

% \textit{(2) Evaluation Metrics.} The final performance of all methods is evaluated using the Intersection over Union (IoU) for each class and the mean Intersection over Union (mIoU), which is computed by averaging the IoU scores over all categories.

\input{tables/syn2sp}

\input{figures/vis_com}

\textit{(2) Implementation Details.} All experiments are conducted on one NVIDIA RTX 3090 GPU and implemented using PyTorch, MinkowskiEngine~\cite{choy20194d}, and LiDAR\_UDA~\cite{yuan2024density} codebase. To ensure fair comparison with previous works, we adopt MinkUNet34 as the backbone. The network is optimized using the Adam optimizer with an initial learning rate of $2.5 \times 10^{-4}$, which is decayed using a polynomial schedule with a power of 0.9. Our model is trained with 100K iterations. We only use XYZ coordinates as the input feature and set the batch size to 2. Following PCAN~\cite{yuan2024density}, scans of the source domain are processed by density-guided translator. We set the voxel size $vs$ to $0.05 \mathrm{m}$ and the voxel expansion ratio $d$ to 3, respectively. $\lambda$ and $\beta$ are set to 0.05 and 0.01 to weight the importance of the distillation and regularization losses.

% $\lambda$ is set to 0.05 to balance the segmentation and distillation losses.

\subsection{Main Results}

To validate the effectiveness of CVKD-UDA, we compare it with recent adversarial-based methods~\cite{tsai2018learning,vu2019advent,wang2020classes,luo2019taking,zheng2020unsupervised,yuan2023prototype} and self-training methods~\cite{saltori2022cosmix,kong2023lasermix,xiao2022polarmix,yuan2024density}. Notably, DGT-ST~\cite{yuan2024density} is a two-stage method combining PCAN and SAC-LM. 
% , is included as a strong baseline.

\textit{(1) SynLiDAR $\rightarrow$ SemanticKITTI (Syn $\rightarrow$ Sk).} 
We report the adaptation results in Table~\ref{tab:Syn2Sk_tab}. CVKD-UDA outperforms the source only by 20.6$\%$ and achieves 41.0$\%$ mIoU. Unlike adversarial training, which often suffers from instability and high computational cost, CVKD-UDA is a self-distillation method that provides a simpler and more efficient solution. CVKD-UDA surpasses all adversarial-based approaches, including PMAN (33.7$\%$) and PCAN (37.0$\%$), which are specifically designed for 3D UDA segmentation, demonstrating the effectiveness of CVKD-UDA in learning transferable features. To further showcase the quality of CVKD-UDA as a warm-up model, we integrate it into SAC-LM (SAC-LM+Ours). It achieves a mIoU of 45.9$\%$, outperforming all existing self-training methods, including CoSMix (29.9$\%$), PolarMix (31.0$\%$), and SAC-LM+PCAN (43.1$\%$). SAC-LM+Ours provides the highest mIoU improvement of 25.5$\%$ over the source only and performs best in 12 out of 19 classes. These results demonstrate the effectiveness of CVKD-UDA in providing a strong warm-up model for self-training. 

Notably, CVKD-UDA is particularly effective for large and structurally coherent categories. For example, it achieves 42.5$\%$ IoU on fence, outperforming the second-best method (21.5$\%$ IoU) by +21.0$\%$ IoU. This improvement is consistent with the design of CVKD. Since fence usually spans a large spatial extent and contains more points, the compressed view with a larger voxel size tends to yield more coherent and less fragmented representations. These structural cues can then be distilled to the standard view, leading to improved segmentation performance. Similar gains are also observed on other large, structurally coherent classes (e.g., road and sidewalk).

For the small or thin-structure categories  (e.g., person and pole), CVKD-UDA also maintains competitive performance. However, all methods still achieve relatively low IoU scores for the truck and the motorcyclist. After taking a closer look at the SemanticKITTI training set, we attribute this to three factors. First, CVKD-UDA involves a trade-off between transferability and discriminability. The compressed view facilitates structural alignment across domains, but inevitably merges fine-grained geometric details, which can be more detrimental to rare and visually ambiguous categories. Second, the truck class exhibits a pronounced appearance gap between synthetic and real scans (e.g., point density and local geometric patterns), making cross-domain adaptation particularly difficult (Please see Sec.~\ref{exp:analysis} for detailed comparisons). This challenge is also reflected in the result of source only in Table~\ref{tab:Syn2Sk_tab}, where the truck IoU is only 0.6$\%$. Third, these categories have limited training samples in the target domain. Although SemanticKITTI contains 19,130 training scans, motorcyclists and trucks appear in only 552 frames ($\approx$87.8K points) and 2.3K frames ($\approx$4.59M points), respectively. Such severe data scarcity limits the model from learning these categories sufficiently.

\textit{(2) SynLiDAR $\rightarrow$ SemanticPOSS (Syn $\rightarrow$ Sp).} We list the results in Table~\ref{tab:Syn2Sp_tab}. CVKD-UDA achieves 47.5$\%$ mIoU, surpassing PCAN (44.4$\%$) and PMAN (46.5$\%$). Notably, PMAN employs a complex multi-task network, which is particularly beneficial under this data-limited scenario. In contrast, CVKD-UDA adopts a simpler framework and obtains improved results. When combined with the self-training methods SAC-LM (SAC-LM+Ours), the performance improves to 51.0$\%$ mIoU, slightly higher than SAC-LM+PCAN (50.8$\%$). CVKD-UDA achieves the best performance in 7 out of 13 categories, demonstrating its effectiveness across different datasets.

In this setting, the per-class results exhibit a trend consistent with Syn $\rightarrow$ Sk. CVKD-UDA tends to benefit structurally coherent categories (e.g., ground, fence and build.), while small and sparse objects (e.g., cone. and bi.cle) achieve lower performance than some competitors. This is also caused by the blurring of fine-grained geometry in the compressed view. This side effect is more severe on SemanticPOSS, whose 40-beam scan is inherently sparser than 64-beam LiDAR scan. Although the proposed D-Adapter and auxiliary imitation classifier alleviate this issue to some extent, there remains room for further improvement. Nevertheless, CVKD-UDA still achieves good performance in this setting.

\textit{(3) Visual comparisons.} In Fig.~\ref{fig:vis_com}, we present error maps for Syn $\rightarrow$ Sk (top) and Syn $\rightarrow$ Sp (bottom), where correct and incorrect predictions are painted in gray and red, respectively. Specifically, we compare CVKD-UDA with PCAN, a strong adversarial-based warm-up model training method, and further visualize the results of SAC-LM, which uses PCAN and CVKD-UDA as its warm-up models. 
Overall, CVKD-UDA produces noticeably fewer error predictions than PCAN. On Syn~$\rightarrow$~Sk, CVKD-UDA exhibits fewer errors on terrain (row 1) as well as on sidewalk and fence (row 2). A similar trend is shown for fence on Syn~$\rightarrow$~Sp. These results demonstrate that our method is particularly effective at segmenting large, structurally coherent categories, resulting in more accurate predictions.

Moreover, using CVKD-UDA as the warm-up model consistently benefits the subsequent self-training. Compared with SAC-LM initialized by PCAN, SAC-LM+CVKD-UDA exhibits fewer error predictions and cleaner boundaries in the highlighted regions (columns 3 and 5). This improvement is also evident for the sidewalk on Syn~$\rightarrow$~Sk and for the fence in Syn~$\rightarrow$~Sp, both of which show substantially fewer errors. These qualitative results confirm that a better warm-up model yields higher quality pseudo-labels, thereby mitigating error accumulation in self-training. However, for small-object categories (e.g., bicyclist), CVKD-UDA still produces some errors. These mistakes introduce noisy pseudo-labels during self-training, leaving room for further improvement.

\subsection{Ablation Studies}
\label{exp:ab_studies}

\textit{(1) CVKD.} We validate the effectiveness of CVKD in Table~\ref{tab:ab_components}, where comparisons are made against the source only and the same-view knowledge distillation (SVKD) variant. CVKD significantly improves performance over the source only, boosting mIoU from 20.4$\%$ to 39.0$\%$ (rows 0 and 1), resulting in an 18.6$\%$ gain. However, replacing CVKD with SVKD in our full framework, the mIoU drops by 14.2$\%$ (rows 6 and 8). These results demonstrate the advantage of transferring knowledge across complementary views and confirm the effectiveness of CVKD in enhancing model generalization across domains.

\textit{(2) Auxiliary Imitation Classifier (AIC).} We validate the effectiveness of using an imitation classifier to perform CVKD in Table~\ref{tab:ab_components}. Adding AIC to CVKD boosts the performance to 40.5$\%$, achieving a 1.5$\%$ improvement (rows 1 and 2). This indicates that decoupling CVKD from the main classifier helps preserve the discriminative capacity of the standard view and benefits for cross-view knowledge transfer.

\input{tables/abCVKD}
\input{tables/place_D_Adapter}
\input{tables/CVKD_warmUp}

\input{tables/mean_teacher}

\input{tables/param_d}

\input{tables/lambda_beta_para.tex}

\textit{(3) D-Adapter (D-A).} Comparing the results in Rows 2 and 4 of Table~\ref{tab:ab_components}, we observe that incorporating the D-Adapter results in a 0.8$\%$ performance drop. However, when serving as a warm-up model for LaserMix (LM), the variant with the D-Adapter outperforms the one without it by 0.8$\%$ (Rows 3 and 5), achieving 42.3$\%$ mIoU. We attribute this to a reduced discriminative capability of the model without the D-Adapter. A less discriminative model limits the effectiveness of self-training. Although the model with the D-Adapter shows slightly lower performance, it achieves a better balance between transferability and discriminability, making it more suitable and reliable for self-training.

\input{sup_figures/tsne_vs}

To gain deeper insights into the effectiveness of D-Adapter itself, as shown in Table~\ref{tab:ab_CVadapters}, we remove the entropy regularization and conduct an ablation study on its placement within the MinkUNet34 architecture. The MinkUNet34 consists of 8 residual blocks. Placing the D-Adapter before the final decoder block (block\_8) yields the second-best result (39.7$\%$), achieving a favorable trade-off between accuracy and computational cost. Placing it in block\_1 yields a similar result (39.6$\%$), only 0.1$\%$ lower than block\_8. Moreover,  we explore whether using multiple D-Adapters at different layers can bring additional performance gains. Employing D-Adapters at both block\_7 and block\_8 leads to a slight improvement of 0.4$\%$, but this comes with increased computational cost. Building on this, we further evaluate a variant that places multiple D-Adapters in the model, i.e., after block\_6, block\_7, and block\_8. However, this setting significantly degrades the final performance, resulting in only 36.6$\%$ mIoU. This is because leveraging too many D-Adapters will reduce the transferability of the model, limiting its ability to generalize well across domains. These results demonstrate that a single, well-placed D-Adapter at a high-level layer (\textit{i.e}., block\_8) of the backbone network is sufficient and efficient.

\textit{(4) CVKD-UDA as the Warm-up Stage.} As shown in Table~\ref{tab:warmUp_CVKD}, taking a two-stage training pipeline that adopts CVKD-UDA as the warm-up stage consistently improves CosMix and LaserMix on Syn $\rightarrow$ Sk and Syn $\rightarrow$ Sp. Specifically, CVKD-UDA brings substantial performance gains, improving the mIoU of CosMix and LaserMix by 13.4$\%$ and 6.8$\%$ on Syn $\rightarrow$ Sk, respectively. Similarly, it boosts the mIoU by 4.0$\%$ for CosMix and 3.6$\%$ for LaserMix on Syn $\rightarrow$ Sp. As shown in Table~\ref{tab:Syn2Sk_tab}, when using the source only model as the warm-up model, CosMix (29.9$\%$) performs worse than LaserMix (36.0$\%$). However, when initialized with CVKD-UDA, CosMix (43.3$\%$) surpasses LaserMix (42.8$\%$) by 0.5\% mIoU. Since CosMix largely depends on pseudo labels to perform cross-domain sample mixing,  the high-quality pseudo labels generated by CVKD-UDA effectively enhance its adaptation capability. These results further emphasize the crucial role of a well-initialized warm-up model in self-training frameworks, and CVKD-UDA can be seamlessly integrated into existing self-training pipelines.

\textit{(5) Mean-Teacher Framework.} Since CVKD-UDA leverages unlabeled target domain data in the warm-up stage, ensuring training stability is critical. To this end, we adopt the Mean-Teacher framework, where the teacher model is updated via the exponential moving average (EMA) of the student model, providing stable guidance for cross-view distillation. As shown in Table~\ref{tab:use_MT}, incorporating the Mean-Teacher framework improves the performance of CVKD-UDA, increasing the mIoU from 38.5$\%$ to 41.0$\%$. These results demonstrate that the Mean-Teacher framework stabilizes the training process and enhances the effectiveness of knowledge transfer between views.

\textit{(6) Voxel Size Expansion Ratio $d$.} In Table~\ref{tab:praram_d}, we evaluate the impact of the voxel size expansion ratio by varying it from 2 to 5. Increasing the ratio from 2 to 3 improves the mIoU from 40.1$\%$ to 41.0$\%$, indicating that moderate spatial compression can reduce domain shift and enhance the model's transferability. However, further increasing the ratio to 4 and 5 loses geometric details necessary to maintain the discriminability, and the performance reduces to 40.2$\%$ and 38.9$\%$, respectively. These results highlight the importance of selecting an appropriate voxel size to balance transferability and discriminability. Setting the ratio to 3 achieves the best trade-off.

% We evaluate the effect of $\lambda$ by varying it from 0.5 to 0.01 in Table~\ref{tab:param_lambda_beta}. 
\textit{(7) Distillation Weight $\lambda$.} We evaluate the effect of $\lambda$ in Table~\ref{tab:param_lambda_beta}. Gradually reducing $\lambda$ improves performance, with the best mIoU of 41.0$\%$ achieved at $\lambda = 0.05$. This indicates that using a large weight to overemphasize the cross-view distillation will hinder the main segmentation task. A smaller $\lambda$ helps the model to get a balance between transferability and discriminability. However, a small $\lambda$ (0.01) slightly drops the mIoU to 40.4$\%$, showing that a moderate balance is important.

\textit{(8) Entropy Regularization Weight $\beta$.} We study the influence of $\beta$ in Eq.~(8) in Table~\ref{tab:param_lambda_beta}. The performance gradually improves as $\beta$ decreases from 0.1 to 0.01, reaching the best mIoU of 41.0$\%$ at $\beta=0.01$. This indicates that assigning a moderate weight to entropy regularization is crucial for stable optimization. A large $\beta$ overemphasizes entropy minimization, causing the model to produce overconfident predictions that hinder cross-view distillation. Conversely, an excessively small $\beta$ weakens the regularization effect and allows uncertain predictions. $\beta=0.01$ provides the best trade-off between regularization strength and discriminative learning.

\subsection{Analysis}
\label{exp:analysis}

\textit{(1) Comparison between the standard and compressed views.} In this work, we discover that moderately increasing the voxel size facilitates the construction of domain-similar voxelized representations, which helps reduce the domain gap at the input level. To illustrate how the compressed view enhances cross-domain transferability, we conduct a detailed comparison between the standard and compressed views from two perspectives: a) the data distributions and b) the learned feature distributions.

a) Data distributions. Previous works~\cite{yi2021complete,li2023adversarially,yuan2024density} highlight that significant differences in point density between the source and target domains are the major cause of the domain gap. Thus, we perform a statistical analysis of each domain and visualize the average number of voxels per scan under different voxel sizes in the above of Fig.~\ref{fig:tsne_vscom}. Specifically, when the voxel size is set to $5cm$, there is a large gap between the two domains. SemanticKITTI has significantly more voxels per scan than SynLiDAR, with a difference of approximately 15000 voxels. However, after moderately increasing the voxel size to $15cm$, the voxel counts in both domains decrease sharply, and the inter-domain gap narrows considerably  (about 2500 voxels). This indicates that the compressed view can help align the data distribution between domains at the input level, thereby mitigating the domain gap.

b) Feature distributions. In Figs.~\ref{fig:tsne_vscom}(c)-(f), we further visualize the learned features of 10 representative categories, covering large-object classes (e.g., road and terrain) and small-object classes (e.g., person and pole). The second row (Figs.~\ref{fig:tsne_vscom}(c) and (d)) shows the features extracted from the source only trained with different voxel sizes, while the third row (Figs.~\ref{fig:tsne_vscom}(e) and (f)) shows the features learned by CVKD-UDA. With a larger voxel size ($vs=15cm$), features of large objects become more compact and domain-aligned. In contrast, features of small objects appear more scattered than those trained with a smaller voxel size ($vs=5cm$). This suggests that the compressed view improves cross-domain alignment, but it sacrifices spatial geometric detail, leading to reduced discriminability, especially for those classes with small size and sparse points. To address this issue, we propose the D-Adapter and an auxiliary imitation classifier that work together to preserve discriminability while retaining the transferability benefits of the compressed view.

Compared with the source only baseline, the feature of CVKD-UDA becomes noticeably structured for both voxel sizes. Under the standard view, clusters of large-object categories (e.g., road, car, terrain, and fence) become more compact and better separated from other classes in Fig.~\ref{fig:tsne_vscom}(e), indicating that CVKD-UDA strengthens semantic coherence while leveraging the transferability cues introduced by the compressed view. For small-object categories (e.g., person, pole, and traffic sign), CVKD-UDA reduces extreme dispersion to some extent, as shown in Fig.~\ref{fig:tsne_vscom}(f). However, their distributions remain relatively diffuse, indicating that small objects remain challenging. Overall, the compressed view shows a consistent trend with the standard view. This makes it a reliable auxiliary supervision signal for CVKD to perform cross-view knowledge transfer, thereby boosting the final performance.

\input{rebuttal_tables/syn2sk_Src_ForPaper.tex}

\textit{(2) Source domain performance.} We report the results of source only, PCAN, and CVKD-UDA on the source domain in Table~\ref{tab:Syn2Sk_src_tab}. Since SynLiDAR is a synthetic dataset and its official release does not provide the test split, these results are reported on the official training split for diagnostic comparison (i.e., to assess how well different methods preserve source-domain discriminability). Overall, the three models have good performance on the source domain, achieving over 90$\%$ mIoU. The two UDA methods (PCAN and CVKD-UDA) exhibit a performance drop, which is a common side effect of enforcing cross-domain alignment (i.e., the trade-off between discriminability and transferability). Although CVKD-UDA shows a slightly larger drop than PCAN (-4.1$\%$ vs. -2.9$\%$), it yields substantial gains in the target domain, improving mIoU by 3.1$\%$ over PCAN on Syn $\rightarrow$ Sk. These results confirm that CVKD-UDA achieves more effective adaptation while still retaining reasonable discriminability.

\begin{table}[t]
  \centering
  \caption{Comparison results of CVKD-UDA trained with different data configurations on Syn $\rightarrow$ Sk.}
  \resizebox{0.79\linewidth}{!}
  {
    \begin{tabular}{@{}ccc|c@{}}
      \toprule
       Method    & Source data & Target data & mIoU          \\
      \midrule
      Source only & \Checkmark  &             & 20.4          \\
      CVKD-UDA    & \Checkmark  &             & 33.9          \\
      CVKD-UDA    &             & \Checkmark  & 3.1           \\
      CVKD-UDA    & \Checkmark  & \Checkmark  & \textbf{41.0} \\
      \bottomrule
    \end{tabular}
  }
\vspace{-0.4cm}
  \label{tab:use_data}
\end{table}

\textit{(3) Training data of CVKD-UDA.} We report the results of CVKD-UDA under three training data settings in Table~\ref{tab:use_data}. When trained solely with the source data, the model is supervised using both the segmentation loss $\mathcal{L}_{ce}^{s}$ and the distillation loss $\mathcal{L}_{distill}^{s}$, achieving an mIoU of 33.9$\%$. This substantial improvement over the source only setting (20.4$\%$) indicates that the model generalizes better to the target domain. In the target only setting, where ground-truth annotations are unavailable, the model is trained solely with the distillation loss $\mathcal{L}_{distill}^{t}$. A significantly lower mIoU of 3.1$\%$ is obtained, demonstrating that the absence of segmentation supervision limits the model’s discriminative capacity and degrades the final performance. Moreover, because the proposed CVKD-UDA is a label-free framework, it can be trained using the source and target domain data without requiring any target annotations. CVKD-UDA achieves the best performance in this setting, with an mIoU of 41.0$\%$, demonstrating that CVKD-UDA enhances the model’s ability to perceive target domain data.

\input{sup_figures/truck_com.tex}

\textit{(4) Large gap of the truck category}. As shown in Fig.~\ref{fig:truck_com}, truck instances exhibit pronounced cross-domain discrepancies between SemanticKITTI and SynLiDAR in shape, point density, and local geometry. First, in terms of shape, truck instances in the synthetic SynLiDAR dataset typically possess idealized, complete, and rigid geometric structures. In contrast, real trucks in SemanticKITTI exhibit much higher morphological diversity and are often partially occluded in complex traffic scenes. Second, point density differs drastically. Synthetic trucks typically exhibit a dense, noise-free point distribution, whereas real-world scans show non-uniform density that decreases with distance, resulting in sparse and irregular point patterns. Finally, regarding local geometry, real-world trucks contain intricate structural details (e.g., external mirrors in 07\_0064) and are affected by sensor noise and beam absorption on reflective materials, which are absent in the synthetic scans. These discrepancies make the truck category inherently difficult for cross-domain knowledge transfer. Nevertheless, CVKD-UDA still improves the truck IoU from 0.6$\%$ (source only) to 8.1$\%$ on Syn $\rightarrow$ Sk, achieving the second-best result among the methods compared in Table~\ref{tab:Syn2Sk_tab}.

\input{sup_figures/tsne_com.tex}

\textit{(5) t-SNE Feature Visualization.} In Fig.~\ref{fig:tsne_com}, we use t-SNE~\cite{van2008visualizing} to visualize features of the target domain learned by different methods. Compared to the adversarial learning method PCAN, our CVKD-UDA produces more compact and well-separated clusters. For instance, the car, fence, and building are more clearly separated. These results demonstrate that CVKD-UDA learns more discriminative and structured representations in the target domain. Moreover, since CVKD-UDA is proposed to produce a well-initialized warm-up model for self-training, we further compare the extracted features that use PCAN or CVKD-UDA as the warm-up stage and SAC-LM as the self-training stage. When using the PCAN model as the warm-up model, the features of the fence and building are difficult to distinguish. In contrast, SAC-LM using CVKD-UDA as the warm-up stage shows a more precise separation. These results demonstrate that CVKD-UDA achieves better performance and can serve as an effective warm-up training method.

%% file: tables/syn2sk.tex
\begin{table*}[t]
    \centering
    \caption{Comparison results of SynLiDAR $\rightarrow$ semanticKITTI adaptation in terms of mIoU. A, D, and S denote adversarial training, knowledge distillation, and self-training. $\Delta$ denotes the relative improvement over the source only.}
    \setlength{\tabcolsep}{2.2pt} % 减小列间距，可以根据需要调整数值
    \resizebox{\linewidth}{!}{
        % \small
        % \large
        \begin{tabular}{@{}l|c|ccccccccccccccccccc|cc@{}}
            \toprule
            Methods                             & \rotatebox{90}{Mech.} & \rotatebox{90}{car} & \rotatebox{90}{bi.cle} & \rotatebox{90}{mt.cle} & \rotatebox{90}{truck} & \rotatebox{90}{oth-v.} & \rotatebox{90}{pers.} & \rotatebox{90}{bi.clst} & \rotatebox{90}{mt.clst} & \rotatebox{90}{road} & \rotatebox{90}{parki.} & \rotatebox{90}{sidew.} & \rotatebox{90}{other-g.} & \rotatebox{90}{build.} & \rotatebox{90}{fence} & \rotatebox{90}{veget.} & \rotatebox{90}{trunk} & \rotatebox{90}{terr.} & \rotatebox{90}{pole} & \rotatebox{90}{traf.} & \rotatebox{90}{mIoU} & $\Delta$ \\
            % \rotatebox{90}{gain} \\
            \midrule
            Source only                         & -                     & 35.9                & 7.5                    & 10.7                   & 0.6                   & 2.9                    & 13.3                  & 44.7                    & 3.4         & 21.8                 & 6.9                    & 29.6                   & 0.0                      & 34.1                   & 7.4                   & 62.9                   & 26.0                  & 35.5                  & 30.3                 & 14.1                  & 20.4                 & +0.0                 \\
            \midrule
            \midrule

            AdaptSegNet~\cite{tsai2018learning} & A                     & 52.1                & 10.8                   & 11.2                   & 2.6                   & 9.6                    & 15.1                  & 35.9                    & 2.6                     & 62.2                 & 10.4                   & 41.3                   & 0.1         & 58.1                   & 17.1                  & 68.0                   & 38.4                  & 38.7                  & 35.9                 & 20.4                  & 27.9                 & +7.5                 \\

            CLAN~\cite{luo2019taking}           & A                     & 51.0                & 15.8                   & 16.8                   & 2.2                   & 7.8                    & 18.7                  & 46.8                    & 3.0                     & 68.9                 & 11.1                   & 44.9                   & 0.1         & 59.6                   & 17.5                  & 71.7                   & 41.1                  & 44.0                  & 37.7     & 19.8                  & 30.5                 & +10.1                \\

            ADVENT~\cite{vu2019advent}          & A                     & 59.9                & 13.8                   & 14.6                   & 3.0                   & 8.0                    & 17.7                  & 45.8                    & 3.0                     & 67.6                 & 11.3       & 45.6                   & 0.1                     & 61.7                   & 15.8                  & 72.4                   & \underline{41.5}      & 47.0      & 34.5                 & 15.3                  & 30.5                 & +10.1                \\

            FADA~\cite{wang2020classes}         & A                     & 49.9                & 6.7                    & 5.1                    & 2.5                   & 10.0                   & 5.7                   & 26.6                    & 2.3                     & 65.8                 & 10.8                   & 37.8                   & 0.1         & 60.3                   & \underline{21.5}      & 60.4                   & 37.2                  & 31.9                  & 35.4                 & 17.4                  & 25.6                 & +5.2                 \\

            MRNet~\cite{zheng2020unsupervised}  & A                     & 49.5                & 11.0                   & 12.2                   & 2.2                   & 8.6                    & 16.0                  & 46.4                    & 2.7                     & 60.0                 & 10.5                   & 41.9                   & 0.1         & 55.1                   & 16.5                  & 68.1                   & 38.0                  & 40.7                  & 36.5                 & 20.8      & 28.3                 & +7.9                 \\

            PMAN~\cite{yuan2023prototype}       & A                     & 71.0                & 14.9                   & 24.8                   & 1.6                   & 6.6                    & 23.6                  & \underline{61.1}        & \textbf{5.5}            & 75.3                 & 10.5                   & \underline{54.1}          & 0.1         & 47.9                   & 17.4                  & 69.6                   & 38.6                  & \textbf{61.5}         & 37.0                 & 18.6                  & 33.7                & +13.3                \\
            PCAN~\cite{yuan2024density}         & A                     & \textbf{85.0}       &\underline{17.5}        & \underline{27.4}       & \textbf{10.4}         & \textbf{11.9}          & \underline{27.5}      & \textbf{63.7}           & 2.6                     & \underline{78.1}     & \underline{13.5}       & 50.1                   & 0.1         & \underline{68.5}       & 20.0                  & \underline{76.2}       & 41.3                  & 45.7                  & \underline{41.0}        & \underline{21.8}         & \underline{37.0}                 & +16.6                \\
           
            CVKD-UDA (Ours)                 & D                     & \underline{83.3}    &\textbf{18.8}           & \textbf{35.1}          & \underline{8.1}       & \underline{11.5}       & \textbf{41.8}         & 60.3                    & \underline{3.3}         & \textbf{80.7}        & \textbf{16.7}          & \textbf{54.8}          & 0.0         & \textbf{73.3}          & \textbf{42.5}         & \textbf{79.2}          & \textbf{46.4}        & \underline{56.3}       & \textbf{43.7}        & \textbf{24.1} & \textbf{41.0}    & +20.6              \\

            \midrule
            \midrule

            CoSMix~\cite{saltori2022cosmix}     & S                     & 56.4                & 10.2                   & 20.8                   & 2.1                   & \textbf{13.0}          & 25.6                  & 41.3                    & 2.2                     & 67.4                 & 8.2                    & 43.4                   & 0.0         & 57.9                   & 12.2                  & 68.4                   & 44.8                  & 35.0                  & 42.1                 & 17.0            & 29.9                 & +9.5                 \\

            PolarMix~\cite{xiao2022polarmix}    & S                     & -                   & -                      & -                      & -                     & -                      & -                     & -                       & -                       & -                    & -                      & -                      & -           & -                      & -                     & -                      & -                     & -                     & -                    & -                & 31.0                 & +10.6                \\

            LaserMix~\cite{kong2023lasermix}    & S                     & 90.3                & 7.8                    & 37.2                   & 2.3                   & 2.4                    & 40.6                  & 49.1                    & \textbf{5.1}            & 80.5                 & 9.9                    & 57.4                   & 0.0         & 57.6                   & 3.4                   & 77.6                   & 46.6                  & \underline{60.1}         & 42.0                 & 13.6             & 36.0                 & +15.6                \\

            SCT~\cite{xiao2022transfer}         & S                     & 81.9                & 3.7                    & 31.2                   & 1.6                   & \underline{7.4}        & 44.6                  & \textbf{61.1}           & 3.4                     & 78.2                 & 3.5                    & 51.2                   & 0.0         & 68.0                   & \underline{31.7}      & 74.3                   & 45.8                  & 51.0                  & 41.7                 & 4.1             & 36.0                 & +15.6                \\

            SAC-LM+PCAN~\cite{yuan2024density}  & S                     & \textbf{92.9}       & \textbf{17.3}          & \underline{43.4}       & \textbf{15.0}         & 6.1                    & \textbf{49.2}         & \underline{54.2}        & \underline{4.2}         & \underline{86.4}     & \underline{19.1}       & \underline{62.3}       & 0.0         & \underline{78.2}      & 9.2                    & \textbf{83.3}          & \underline{56.0}      & 59.1                  & \underline{51.2}    & \underline{32.3}  & \underline{43.1}     & +  22.7              \\
           
            SAC-LM+Ours                    & S                     & \underline{92.2}    & \underline{16.6}       & \textbf{44.9}          & \textbf{15.0}              & 5.6                    & \underline{47.8}      & 54.0                    & 3.1                     & \textbf{86.7}        & \textbf{20.3}          & \textbf{66.2}          & 0.0         & \textbf{81.4}         & \textbf{51.2}          & \underline{80.9}       & \textbf{58.7}         & \textbf{62.5}         & \textbf{51.5}       & \textbf{32.9}     & \textbf{45.9}      & +25.5              \\

            \bottomrule
        \end{tabular}
    }
    
    \label{tab:Syn2Sk_tab}
\end{table*}

%% file: tables/syn2sp.tex
%%%%%    Syn2SP  XYZ Table %%%%%%
\begin{table*}[h]
        \centering
        \caption{Comparison results of SynLiDAR $\rightarrow$ semanticPOSS adaptation in terms of mIoU. A, D, and S denote adversarial training, knowledge distillation, and self-training. $\Delta$ denotes the relative improvement over the source only.}
        \setlength{\tabcolsep}{3.5pt} % 减小列间距，可以根据需要调整数值
        \resizebox{\linewidth}{!}{
            \small
            % \large
            \begin{tabular}{@{}l|c|ccccccccccccc|cc@{}}
                \toprule
    
                Methods                               & Mech. & bi.clst          & car              & trunk            & veget.           & traf.            & pole             & garb.            & build.           & cone.            & fence            & bi.cle           & ground           & pers.            & mIoU             & $\Delta$  \\

                \midrule
    
                Source only                           & -     & 47.2             & 43.6             & 37.8             & 70.3             & 11.1             & 33.8             & 19.5             & 67.9             & 11.2             & 19.9             & 9.6              & 77.9             & 47.8             & 38.3             & +0.0  \\
                
                \midrule
                \midrule
                
                AdaptSegNet ~\cite{tsai2018learning}  & A     & 43.9             & 48.2             & 39.0             & 69.6             & 15.5             & 33.6             & 21.3             & 64.3             & 12.7             & 25.0             & 11.6             & 76.0             & 49.9             & 39.3             & +1.0  \\
    
                CLAN    ~\cite{luo2019taking}         & A     & 43.9             & 46.6             & \underline{41.3} & 71.0             & 15.1             & 34.3             & 20.4             & 69.6             & 9.5              & 23.2             & 12.0             & 75.1             & 51.3             & 39.5             & +1.2  \\
    
                ADVENT~\cite{vu2019advent}            & A     & 44.6             & 47.6             & 40.3             & 71.2             & 15.6             & 35.6             & 22.0             & 68.4             & 10.6             & 25.9             & 10.4             & 76.7             & 52.3             & 40.1             & +1.8  \\

                FADA ~\cite{wang2020classes}          & A     & 39.6             & 41.2             & 38.8             & 69.2             & 16.3             & 32.1             & 18.1             & 67.9             & 11.5             & 22.0             & \underline{13.0} & 71.4             & 47.9             & 37.6             & -0.7  \\
    
                MRNet   ~\cite{zheng2020unsupervised} & A     & 43.5             & 47.2             & 39.1             & 70.4             & 15.5             & 32.8             & 22.0             & 66.1             & \underline{13.2} & 24.2             & 11.2             & 76.8             & 50.0             & 39.4             & +1.1  \\
    
                PMAN~\cite{yuan2023prototype}         & A     & \underline{52.6} & 61.5         & \textbf{46.8}        & \textbf{75.1}    & 18.8             & \textbf{36.5}    & 21.4             & 74.7             & \textbf{18.3}    & 25.8             & \textbf{37.5}    & 73.7             & \underline{61.9} & \underline{46.5} & +8.2  \\
    
                PCAN~\cite{yuan2024density}           & A     & 48.6             & \underline{62.1} & 37.5             & 74.0             & \textbf{23.9}    & 31.4             & \underline{22.2} & \underline{76.9} & 6.5              & \underline{41.9} & 11.9             & \underline{79.1} & 61.2             & 44.4             & +6.1  \\

                CVKD-UDA (Ours)                   & D     & \textbf{53.0}    & \textbf{66.1}    & 38.2             & \underline{74.8} & \underline{19.4} & \underline{36.3} & \textbf{39.6}    & \textbf{80.6}    & 10.6             & \textbf{49.0}    & 6.2              & \textbf{81.1}    & \textbf{62.3}    & \textbf{47.5}    & +9.2  \\

                % & +6.1  \\

                \midrule
                \midrule
                CoSMix~\cite{saltori2022cosmix}       & S     & 53.6             & 47.6             & 44.8             & \underline{75.1}    & 16.8             & \underline{37.9} & 25.3             & 72.7             & \underline{19.9} & 39.7              & \underline{10.8} & \underline{80.0}    & 56.5             & 44.6             & +6.3  \\
                PolarMix~\cite{xiao2022polarmix}      & S     & -                & -                & -                & -                   & -                & -                & -                & -                & -                & -                 & -                & -                   & -                & 30.4             & -8.3  \\
                LaserMix~\cite{kong2023lasermix}      & S     & \textbf{58.4}    & 61.3             & \underline{47.7} & 69.0                & 21.9             & \textbf{39.5}    & 30.9             & 61.0             & 16.1             & 36.5              & 7.1              & 79.5                & 62.6             & 45.5             & +7.2  \\
               
                SCT~\cite{xiao2022transfer}           & S     &54.3              & 24.5             & \textbf{52.3}    & 62.1                & \textbf{40.3}    & 37.6             & 2.5              & 69.7             & \textbf{31.7}    & 42.7              & \textbf{47.6}    & 79.5                & 57.1             & 46.3             & +8.0  \\

                SAC-LM+PCAN~\cite{yuan2024density}    & S     & \underline{55.1} & \textbf{70.7}    & 46.1             & 74.2                & 30.1             & 36.3             & \underline{44.1} & \underline{81.0} & 4.3              & \textbf{62.8}     & 10.3             & 78.5                & \underline{67.2} & \underline{50.8} & +12.5 \\
               
                SAC-LM+Ours                           & S     & \underline{55.1} & \underline{66.6} & 44.0             & \textbf{75.4}       & \underline{30.3} & 35.2             & \textbf{47.9}    & \textbf{82.3}    & 6.3              & \textbf{62.8}    & 8.6               & \textbf{81.2}       & \textbf{67.6}    & \textbf{51.0}    & +12.7 \\

                \bottomrule
            \end{tabular}
        }
        
        \label{tab:Syn2Sp_tab}
    \end{table*}

%% file: figures/vis_com.tex
\begin{figure*}[t]
\centering
\includegraphics[width=1\linewidth]{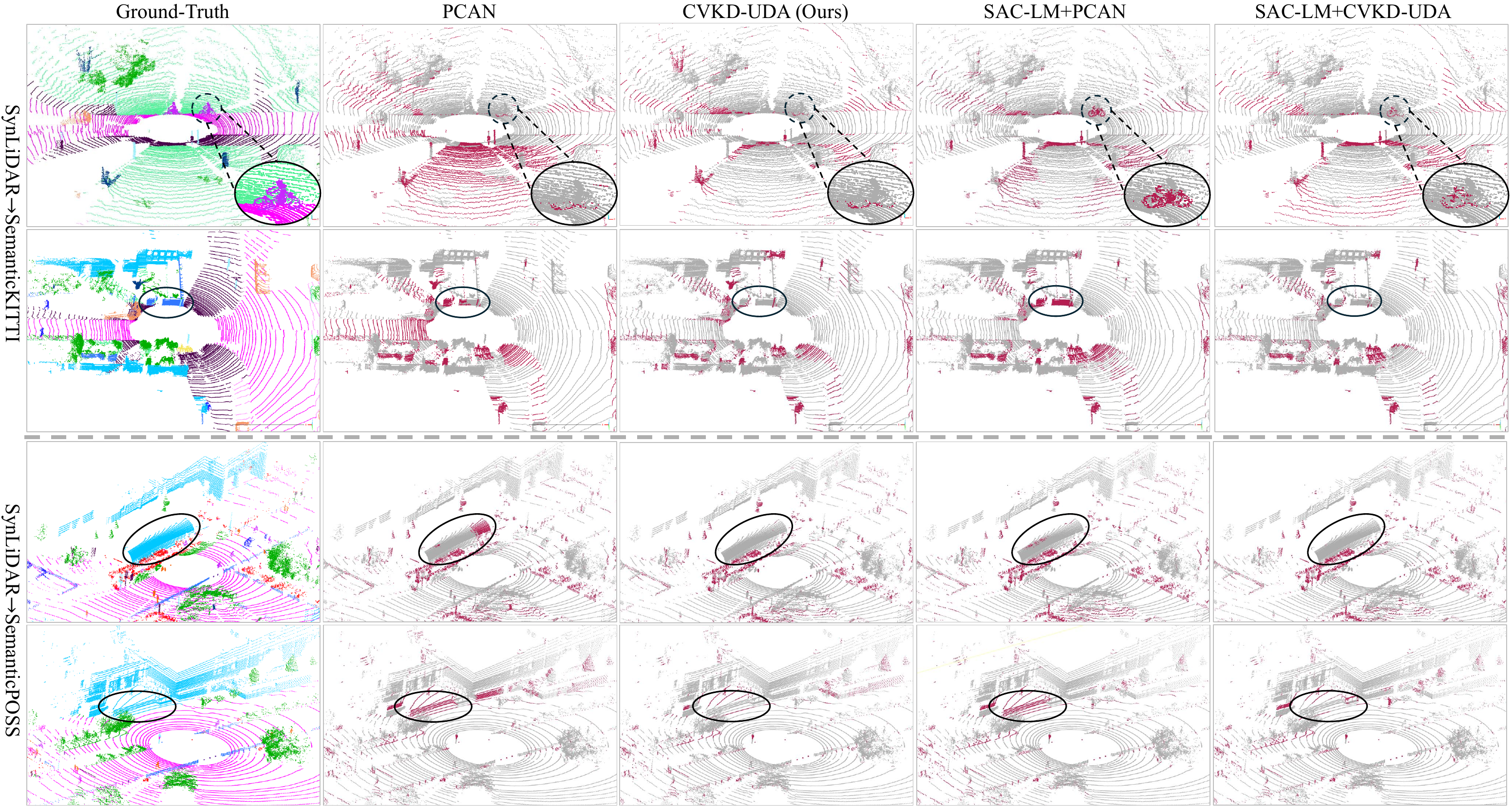}
\caption{Qualitative results (error map) on Syn $\rightarrow$ Sk and Syn $\rightarrow$ Sp. Correct and incorrect predictions are painted in gray and red to emphasize the differences between methods. Black circles highlight regions of interest. }
\vspace{-0.4cm}
\label{fig:vis_com}
\end{figure*}

%% file: tables/abCVKD.tex
\begin{table}[t]
    % \begin{minipage}[t]{0.47\textwidth}
        \centering
        \caption{Ablation studies of CVKD-UDA on Syn $\rightarrow$ Sk.}
        \setlength{\tabcolsep}{3.2pt}
        \resizebox{\linewidth}{!}{
        \small
        \begin{tabular}{c|c|cccccc|cc}
            \toprule
             &                                               & CVKD       & AIC        & D-A           & $\mathcal{L}^{t}_{\mathrm{ent}}$ & LM         & SVKD       & mIoU          & $\Delta$ \\
             \midrule
            0 &\multirow{7}{*}{\rotatebox{90}{Source only}}  &            &            &               &                                  &            &            & 20.4          & +0.0 \\
            1 &                                              & \checkmark &            &               &                                  &            &            & 39.0          & +18.6 \\
            2 &                                              & \checkmark & \checkmark &               &                                  &            &            & 40.5          & +20.1 \\
            3 &                                              & \checkmark & \checkmark &               &                                  & \checkmark &            & 41.5          & +21.1 \\
            4 &                                              & \checkmark & \checkmark & \checkmark    &                                  &            &            & 39.7          & +19.3 \\
            5 &                                              & \checkmark & \checkmark & \checkmark    &                                  & \checkmark &            & 42.3          & +21.9 \\
            6 &                                              & \checkmark & \checkmark &  \checkmark   & \checkmark                       &            &            & 41.0          & +20.6 \\
            7 &                                              & \checkmark & \checkmark & \checkmark    & \checkmark                       & \checkmark &            & \textbf{42.8} & +22.4 \\
            \midrule
            8 &                                              &            & \checkmark & \checkmark    & \checkmark                       &            & \checkmark & 26.8          & +6.4 \\

            \bottomrule
        \end{tabular}
        }
        \label{tab:ab_components}
    % \end{minipage}%
 \end{table}
 

%% file: tables/place_D_Adapter.tex
\begin{table}[t]
    % \begin{minipage}[t]{0.48\textwidth}
        \centering
        \caption{The results of D-Adapter variants on Syn $\rightarrow$ Sk.}
        \setlength{\tabcolsep}{8pt}
        % \resizebox{\linewidth}{!}{
        \small
        \begin{tabular}{cccc|c}
            \toprule
            block\_8 & block\_7 & block\_6 & block\_1 & mIoU \\
            \midrule
            \checkmark &  &  &  & 39.7 \\
             &  &  & \checkmark & 39.6 \\
            \checkmark & \checkmark &  &  & \textbf{40.0} \\
            \checkmark & \checkmark & \checkmark &  & 36.6 \\
            \bottomrule
        \end{tabular}
        % }
         \label{tab:ab_CVadapters}
\end{table}

%% file: tables/CVKD_warmUp.tex
\begin{table}[t]
        \centering
        \caption{Results comparison of different methods using CVKD-UDA as the warm-up model on Syn $\rightarrow$ Sk and Syn $\rightarrow$ Sp.}
        \setlength{\tabcolsep}{9.5pt}
        \resizebox{\linewidth}{!}{
        \small
        \begin{tabular}{@{}l|cc|cc@{}}
            \toprule
            
            & \multicolumn{2}{c}{Syn$\rightarrow$Sk} & \multicolumn{2}{c}{Syn$\rightarrow$Sp} \\
    \midrule
        Method            & mIoU               & $\Delta$              & mIoU               & $\Delta$              \\
    \midrule
        CVKD-UDA (Ours)   & 41.0               &                       & 45.8               &                   \\
    \midrule
    \midrule
        CosMix            & 29.9               &                       & 44.6               &                   \\
        CosMix+CVKD-UDA   & \textbf{43.3}      & +13.4                 & \textbf{48.6}      &    +4.0           \\
    \midrule
    \midrule
        LaserMix          & 36.0               &                       & 45.5               &                   \\
        LaserMix+CVKD-UDA & \textbf{42.8}      & +6.8                  & \textbf{49.1}      &    +3.6           \\

    \bottomrule
        \end{tabular}
        }
         \label{tab:warmUp_CVKD}
\vspace{-0.4cm}
\end{table}

%% file: tables/mean_teacher.tex
\begin{table}[t]
  \centering
  \caption{Impact of the Mean-Teacher framework on Syn $\rightarrow$ Sk.}
  \small
%   \resizebox{0.7\linewidth}{!}
%   {
    % \setlength{\tabcolsep}{2pt}
    \begin{tabular}{@{}c|c@{}}
      \toprule
      Method                     & mIoU \\
      \midrule
      CVKD-UDA w/o Mean-Teacher  & 38.5 \\
      CVKD-UDA with Mean-Teacher & \textbf{41.0} \\
      \bottomrule
    \end{tabular}
%   }
  \label{tab:use_MT}
\end{table}

%% file: tables/param_d.tex
\begin{table}[t]
    \centering
    \caption{The effect of hyperparameter $d$ on Syn $\rightarrow$ Sk.}
    \setlength{\tabcolsep}{8pt}
    % \resizebox{\linewidth}{!}{
    \small
    \begin{tabular}{c|cccc}
        \toprule
        
        $d$   & 2    & 3              & 4    & 5 \\
        \midrule
        mIoU  & 40.1 & \textbf{41.0} & 40.2 & 38.9 \\
        \bottomrule
    \end{tabular}
    % }
     \label{tab:praram_d}
\end{table}

%% file: tables/lambda_beta_para.tex
\begin{table}[t]
    \centering
    \caption{The effect of hyperparameters $\lambda$ and $\beta$ on Syn $\rightarrow$ Sk.}
    \setlength{\tabcolsep}{2.8pt}
    \small
    \begin{tabular}{c|ccccc|ccccc}
        \toprule
              & \multicolumn{5}{c|}{$\lambda$} & \multicolumn{5}{c}{$\beta$} \\
        \midrule
        Value & 1    & 0.5  & 0.1  & 0.05          & 0.01 &  0.1 & 0.05 & 0.01 & 0.005 & 0.001 \\
        mIoU  & 35.4 & 36.6 & 40.7 & \textbf{41.0} & 40.4 &  40.5 & 40.8 & \textbf{41.0} & 40.6 & 39.9  \\
        \bottomrule
    \end{tabular}
    
     \label{tab:param_lambda_beta}
\end{table}

%% file: sup_figures/tsne_vs.tex
\begin{figure}[t]
\centering
\includegraphics[width=1\linewidth]{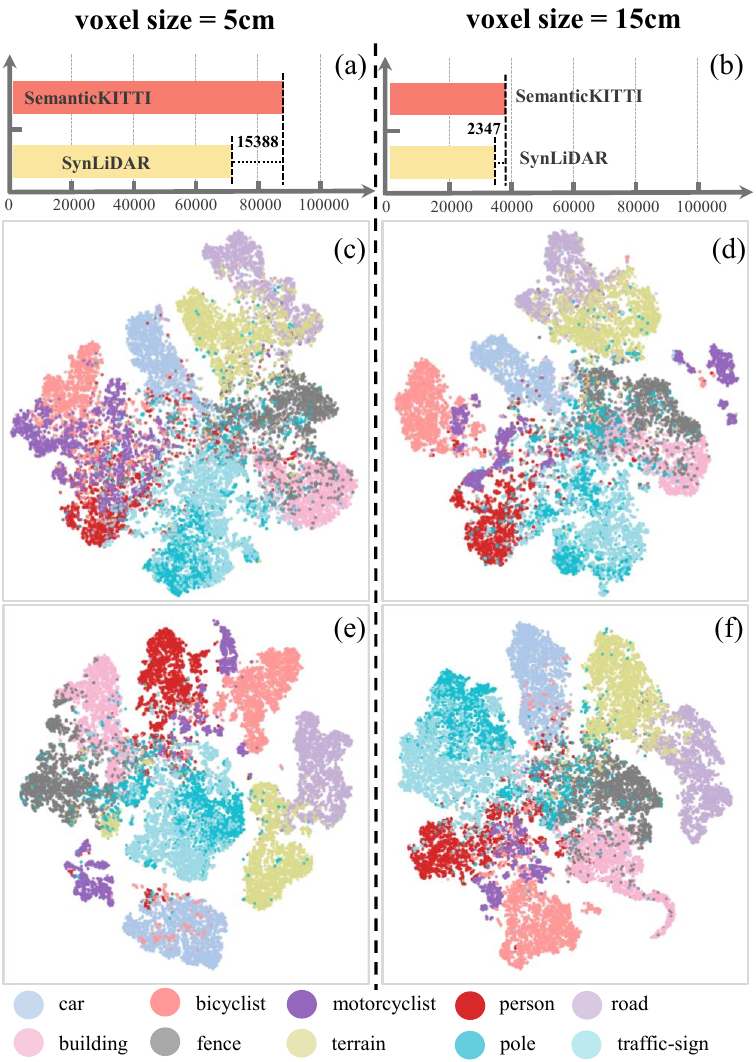}

% \caption{Comparison between the standard and compressed views on Syn $\rightarrow$ Sk. The top row presents the average number of voxels per scan using different voxel sizes across domains, while the bottom row illustrates the t-SNE visualization of features learned by the source only model. For clarity, we only selected 10 representative categories for visualization.}
\caption{Comparison between the standard and compressed views on Syn $\rightarrow$ Sk. (a)–(b) show the average number of voxels per scan under different voxel sizes across domains. (c)–(d) and (e)–(f) visualize the t-SNE visualization of features learned by source only and CVKD-UDA, respectively. For clarity, we only selected 10 representative categories for visualization.}
\label{fig:tsne_vscom}
\end{figure}

%% file: rebuttal_tables/syn2sk_Src_ForPaper.tex
\begin{table}[t]
    \centering
    \caption{Comparison results of the source domain on Syn $\rightarrow$ Sk.}
    % Comparison results of SynLiDAR $\rightarrow$ semanticKITTI adaptation in terms of mIoU.
    %
    % Source-domain (SynLiDAR) results of Syn $\rightarrow$ Sk setting, measured by mIoU.
    \resizebox{0.8\linewidth}{!}{
        \begin{tabular}{c|ccc}
            \toprule
                        & Source only     & PCAN & CVKD-UDA \\
            \midrule
            mIoU        & \textbf{96.7}   & 93.8 & 92.6\\
            $\Delta$    & +0.0            & -2.9 & -4.1 \\
            \bottomrule
        \end{tabular}
    }
    \label{tab:Syn2Sk_src_tab}
\end{table}

%% file: sup_figures/truck_com.tex
\begin{figure}[t]
\centering
\includegraphics[width=1\linewidth]{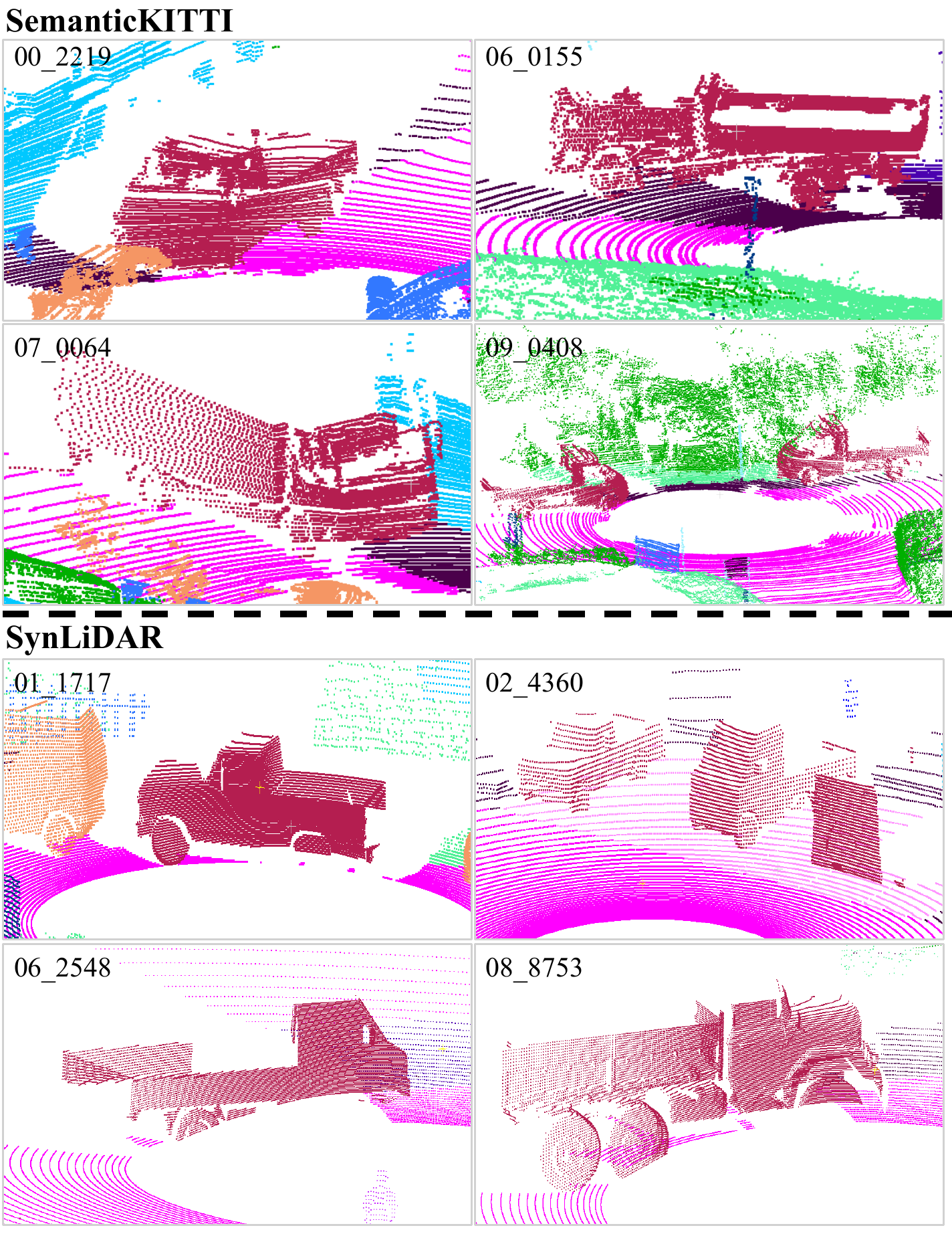}
\caption{Visual comparison of truck samples from SemanticKITTI (top) and SynLiDAR (bottom). The truck exhibits clear differences in shape, point density, and local geometric patterns, indicating a severe domain gap. In each subfigure, the notation sequence\_scan indicates the sequence ID (before the underscore) and the scan ID (after the underscore). }
\label{fig:truck_com}
\end{figure}

%% file: sup_figures/tsne_com.tex
\begin{figure}[t]
\centering
\includegraphics[width=1\linewidth]{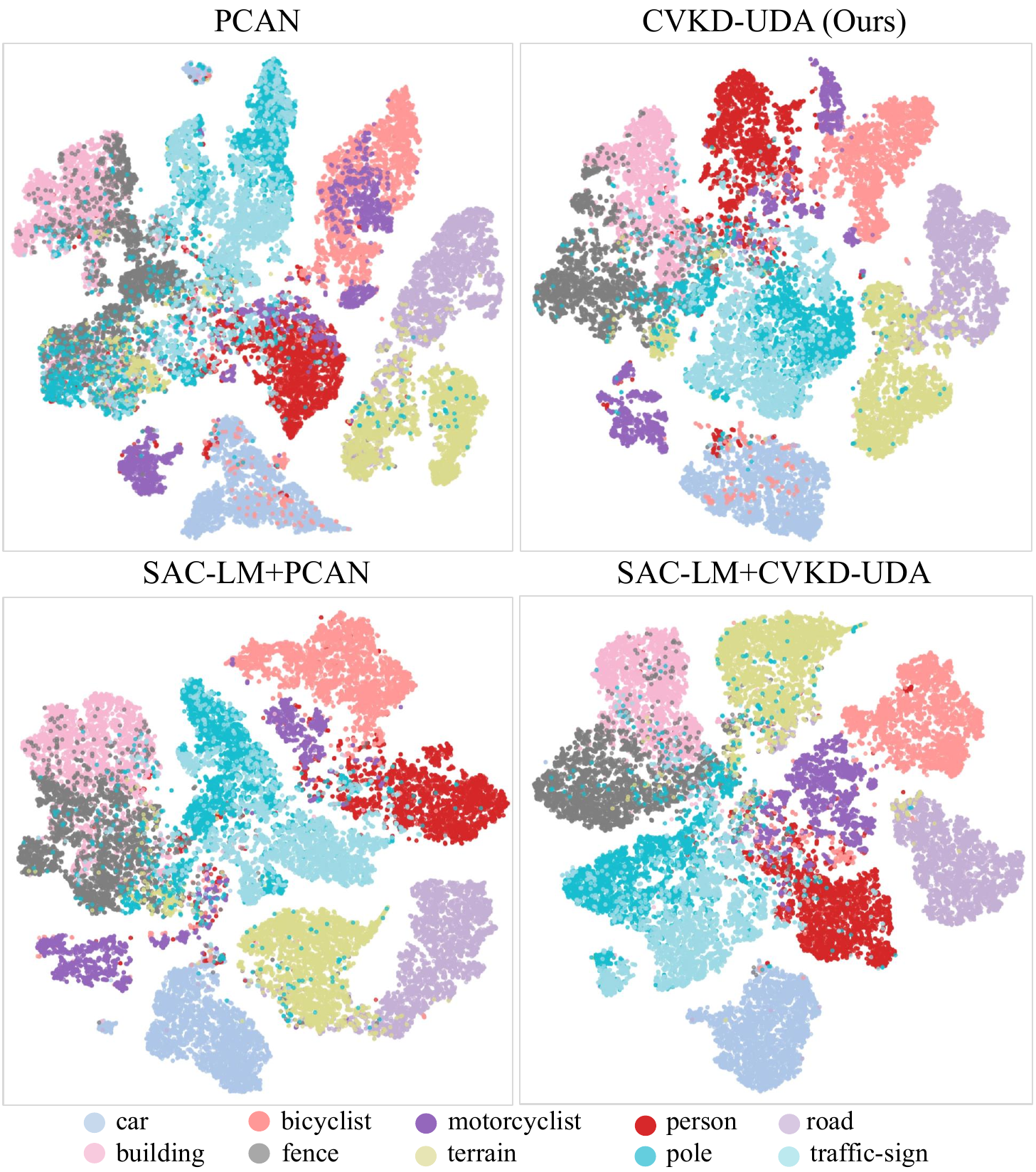}
\caption{t-SNE comparison of features learned by different methods on Syn $\rightarrow$ Sk. For clarity, we only selected 10 representative categories for visualization.}
\label{fig:tsne_com}
\end{figure}

%% file: sections/05_Conclusion.tex
\section{Conclusion}

In this paper, we present CVKD-UDA, a novel framework for training a well-initialized warm-up model by leveraging the critical role of voxel size in cross-domain transferability and discriminability. We construct a compressed view to enhance transferability through higher cross-domain similarity, and a standard view to preserve fine-grained geometry for discriminability. By leveraging both, the proposed CVKD improves the model's generalization and target perception. Moreover, to mitigate interference from the compressed view, we introduce the D-Adapter, auxiliary classifier and the entropy regularization loss, which work synergistically to balance transferability and discriminability. Experiments on two 3D UDA benchmarks show that CVKD-UDA achieves superior performance compared to the adversarial-based methods. We hope our work inspires further research on robust and transferable feature learning for 3D point cloud UDA segmentation.